\pdfoutput=1

\documentclass[11pt]{article}

\usepackage[preprint]{acl}

\usepackage{times}
\usepackage{latexsym}
\usepackage{amssymb}
\usepackage{pifont}
\usepackage{amsmath}
\usepackage{tikz}
\usepackage{listings}
\usepackage{booktabs}
\usetikzlibrary{arrows.meta, positioning}
\usepackage[T1]{fontenc}

\usepackage[utf8]{inputenc}
\usepackage{tipa}
\usepackage{microtype}

\usepackage{inconsolata}

\usepackage{graphicx}
\usepackage{hyperref}
\usepackage{url}
\usepackage{booktabs}
\usepackage{tipa}
\usepackage{multirow}
\usepackage[table]{xcolor}
\usepackage{arydshln}
\usepackage[skins]{tcolorbox} 
\tcbuselibrary{breakable}
\definecolor{lightblue}{RGB}{173, 216, 230}
\usepackage{soul}        
\usepackage{xcolor}
\usepackage{caption}

\title{PunGraph: Retrieval-Enhanced Phonetic-Semantic Graph Reasoning for Pun Understanding}

\author{
 \textbf{Yuchen Su\textsuperscript{1}},
 \textbf{Zijian Huang\textsuperscript{1}},
 \textbf{Yaotian Shi\textsuperscript{1}},
 \textbf{Shaoxin Zhong\textsuperscript{1}},
 \textbf{Ruofan Wang\textsuperscript{1}},
 \\
 \textbf{Mengze Li\textsuperscript{1}},
 \textbf{Yonghua Zhu\textsuperscript{2}\thanks{Corresponding author}},
 \textbf{Diana Benavides-Prado\textsuperscript{3}},
 \textbf{Michael Witbrock\textsuperscript{1}},
\\
 \textsuperscript{1}School of Computer Science, University of Auckland, New Zealand
 \\
 \textsuperscript{2}School of Computer and Information Technology, Shanxi University
 \\
 \textsuperscript{3}School of Electronic Engineering and Computer Science, Queen Mary University of London
 \\
 \texttt{\{ysu132,zhua764\}@aucklanduni.ac.nz, zhuyonghua@sxu.edu.cn}
 \\
 \texttt{d.benavidesprado@qmul.ac.uk, m.witbrock@auckland.ac.nz}
}

\begin{document}
\maketitle
\begin{abstract}
Puns are a challenging form of figurative language that exploit phonetic similarity and semantic ambiguity to convey multiple meanings. Although large language models (LLMs) demonstrate strong language understanding capabilities, they still struggle with pun reasoning due to limited phonetic modeling and uncontrolled end-to-end generation. We propose \textbf{PunGraph}, a retrieval-enhanced knowledge graph framework for pun understanding. PunGraph constructs a phonetic-semantic lexical graph using the Unisyn phonetic dictionary, IPA and G2P representations, and WordNet definitions, and retrieves candidate words or senses to constrain LLM reasoning within a structured candidate space. We further introduce \textbf{WebPun}, a new large-scale dataset containing 5,730 annotated heterographic and homographic puns. Experiments on SemEval-2017 and WebPun show that PunGraph consistently improves the performance of small-scale LLMs and achieves competitive results against strong proprietary models. Further analysis shows that retrieval-guided phonetic and semantic constraints effectively reduce common reasoning errors in pun interpretation, highlighting the benefits of integrating structured knowledge with LLMs. We release our code and dataset at \url{https://github.com/ysu132/PunGraph}.
\end{abstract}

\section{Introduction}

\begin{figure}[h] 
\resizebox{\columnwidth}{!}{
\captionsetup{skip=6pt}
\includegraphics[width=1.1\columnwidth,height=0.6\columnwidth, trim=0 0 0 0, clip]{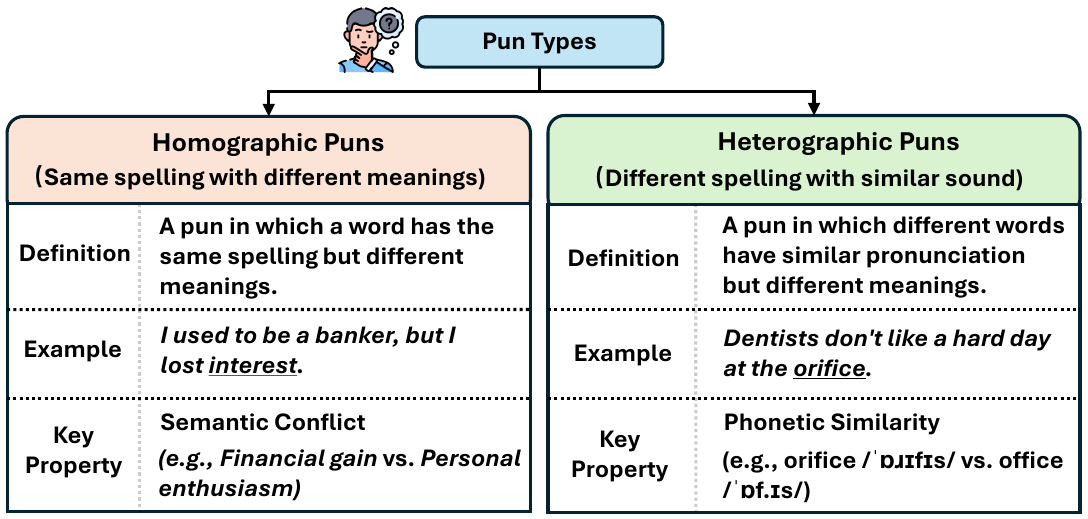}
}
\caption{The types of puns.}
\label{types}
\end{figure}

Puns are a linguistic phenomenon that exploit lexical polysemy or phonetic similarity to evoke multiple meanings within a single utterance, thereby creating a humorous effect \cite{partington2009linguistic, kao2016computational}. As illustrated in Figure~\ref{types}, puns are generally categorized into two main types \cite{xu2024good}: homographic puns and heterographic puns, corresponding to semantic ambiguity and phonetic similarity, respectively. This effect arises from the interaction between phonological resemblance and contextual semantic reasoning, resulting in a humorous interpretation \cite{attardo2018universals}.

Pun reasoning plays a central role in computational humor understanding and remains an important challenge in natural language processing \cite{kao2016computational}. It underlies a broad range of pun-related tasks, including detection \cite{miller2017semeval,zou2019joint}, generation \cite{yu2018neural,sun2022context,tian2022unified}, and interpretation \cite{prnjak2023clef,zangari2025pun}. Among these, pun reasoning serves as a critical intermediate step, requiring accurate semantic analysis of the pun word to enable coherent interpretation of the entire sentence. While LLMs \cite{hurst2024gpt} have demonstrated strong reasoning capabilities across a wide range of NLP tasks \cite{liu2023towards}, they remain limited in handling complex linguistic phenomena such as puns \cite{xu2024good,sravanthi2024pub,mi2025rolling}. This limitation is largely attributed to the scarcity of high-quality training data and the difficulty of integrating multimodal cues (e.g., phonetic information) \cite{su2025survey}, which are essential for capturing both phonological similarity and implicit semantic shifts.

While state-of-the-art proprietary LLMs exhibit some capability in processing humor, open-source small-scale LLMs face exacerbated challenges when tasked with pun reasoning. Constrained by their limited parameter capacity and the scale of training data, these smaller models struggle to internally map the complex interactions between orthography, phonology, and polysemy without explicit structural guidance. Based on our empirical analysis, the severe degradation of small-scale LLMs in pun comprehension primarily stems from two core limitations: (1) \textit{phonological reasoning deficiency}, where models struggle to accurately capture the phonetic relationship between a pun word and its latent alternative word; and (2) \textit{unconstrained semantic generation}, where generated interpretations tend to deviate from the linguistic structure and intended semantic space of the pun.

To address these limitations, we propose \textbf{PunGraph}, a retrieval-enhanced knowledge graph framework for pun understanding. PunGraph constructs a structured lexical knowledge graph that explicitly models phonological associations and semantic relationships between words. This provides critical external knowledge to support pun reasoning, thereby addressing the phonological reasoning deficiency (\textit{Limitation 1}). Unlike conventional end-to-end approaches, PunGraph introduces a retrieval-guided selection mechanism that retrieves candidate words or sense interpretations from the graph. By presenting these as explicit reasoning options, it guides model reasoning within a constrained candidate space, effectively mitigating the issue of unconstrained semantic generation (\textit{Limitation 2}). In addition, to address the scarcity of
high-quality training data, we introduce \textbf{WebPun}, a new large-scale pun dataset collected and annotated from publicly available pun websites, enriching existing benchmarks with more diverse and up-to-date examples. We evaluate PunGraph on both public pun benchmarks and WebPun, and experimental results show that it consistently outperforms strong baselines and achieves competitive performance against large-scale models. 
In summary, our contributions are:
\begin{itemize}
    \item We analyze the limitations of current small-scale models in pun reasoning tasks and explore reasons for these limitations in the context of both heterographic and homographic pun reasoning.
    \item We propose PunGraph, the first knowledge graph reasoning-enhanced LLM framework for pun understanding tasks, promoting better reasoning accuracy based on the actual meaning of pun words.
    \item We construct a new pun reasoning dataset to supplement previously existing resources on pun understanding and further promote community interest in pun tasks.
\end{itemize}

\section{Problem Analysis}
The input to our system consists of a pun sentence $p$ and its corresponding pun word $w_p$. 
\begin{equation}
\begin{aligned}
p = \{ w_1, w_2,  \ldots, w_p, \ldots, w_n \}
\end{aligned}
\end{equation}
where $w_n$ represents the words of the pun sentence. For heterographic puns,
the task objective is to generate an alternative word $w_a$ that shares the same or similar pronunciation as the pun word while conveying a different meaning. In contrast, for homographic puns, the goal is to infer multiple senses of the same word within a given context, which can be represented as $[s_1,s_2,...,s_n]$. For simplicity, we restrict our formulation to the case where the pun involves two senses $[s_1,s_2]$.


Existing LLMs, particularly small-scale models, struggle to capture the complex interplay between phonology and multiple semantic senses required for pun comprehension. To quantify this limitation, we systematically evaluate the performance of current LLMs on pun reasoning tasks.

\textbf{For heterographic puns}, we examine the ability of small-scale models, using Qwen-2.5-7B \cite{qwen2.5} as a representative case study, to directly generate alternative words. As formulated below, performance is evaluated based on the International Phonetic Alphabet (IPA) \cite{ipa1999handbook} similarity score between the original pun word and the generated alternative word:
\begin{equation}
S_{ipa}(p_w,w_a) = 1- \frac{d_{edit}(\phi(p_w),\phi(w_a))}{\max(|\phi(p_w)|,|\phi(w_a)|)}
\end{equation}
where $\phi(\cdot)$ denotes the IPA phonetic representation of a word, and $d_{\text{edit}}$ denotes the edit distance between two phonetic sequences. We define the similarity threshold as $\tau$. Furthermore, we define the number of erroneous samples whose similarity scores are below the threshold as $N_{<}(\tau)$, and the number of erroneous samples whose similarity scores are greater than or equal to the threshold as $N_{\geq}(\tau)$, as formulated below.
\begin{equation}
N_{<}(\tau)
=
\sum_{w_a \in W}
\mathbb{I}\big(S_{ipa}(p_w, w_a) < \tau\big)
\end{equation}
\begin{equation}
N_{\geq}(\tau)
=
\sum_{w_a \in W}
\mathbb{I}\big(S_{ipa}(p_w, w_a) \geq \tau\big)
\end{equation}
where $\mathbb{I}(\cdot)$ is the indicator function. 

As shown in Figure 2, as the threshold $\tau$ increases, $N_{<}(\tau)$ steadily rises, while $N_{\geq}(\tau)$ correspondingly decreases. This trend indicates that a substantial proportion of erroneous predictions generated by LLMs have low phonological similarity to the target pun words, suggesting that the predicted alternative words often deviate considerably from the phonological form required by the original pun. We thereby have:

\noindent \textbf{Limitation 1}: LLMs lack explicit constraints on phonetic similarity during the reasoning process, causing the generated alternative words to fail to satisfy the fundamental phonetic requirements of heterographic puns.

\begin{figure}[t] 
\resizebox{\columnwidth}{!}{
\includegraphics[width=\columnwidth]{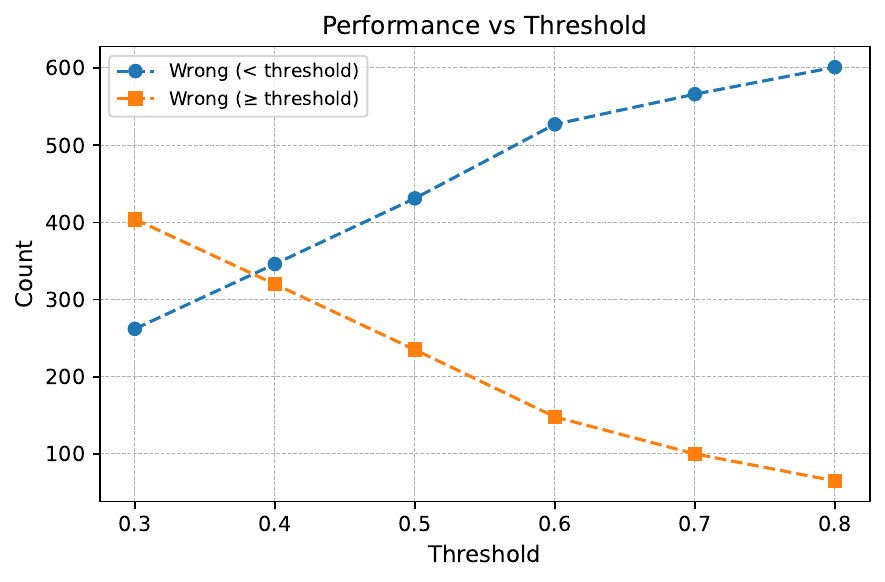}
}
\caption{Performance of Qwen2.5-7B on the heterographic pun replacement task under varying IPA similarity thresholds. The x-axis represents the IPA similarity threshold, and the y-axis denotes the number of erroneous samples. The increase in erroneous samples below the threshold suggests that most generation errors stem from insufficient phonetic similarity. }
\label{performance}
\end{figure}

\textbf{For homographic puns}, we analyze the ability of the open-source model to generate polysemous explanations of synonyms. We define the set of word sense explanations generated by the model as $G$, and the set of dictionary sense definitions as $D$, as shown below.
\begin{equation}
G(p_w)=\{s_1,s_2\}
\end{equation}
\begin{equation}
D(p_w)=\{d_1,d_2,\cdots,d_m\}
\end{equation}

\begin{figure}[t] 
\resizebox{\columnwidth}{!}{
\includegraphics[width=\columnwidth]{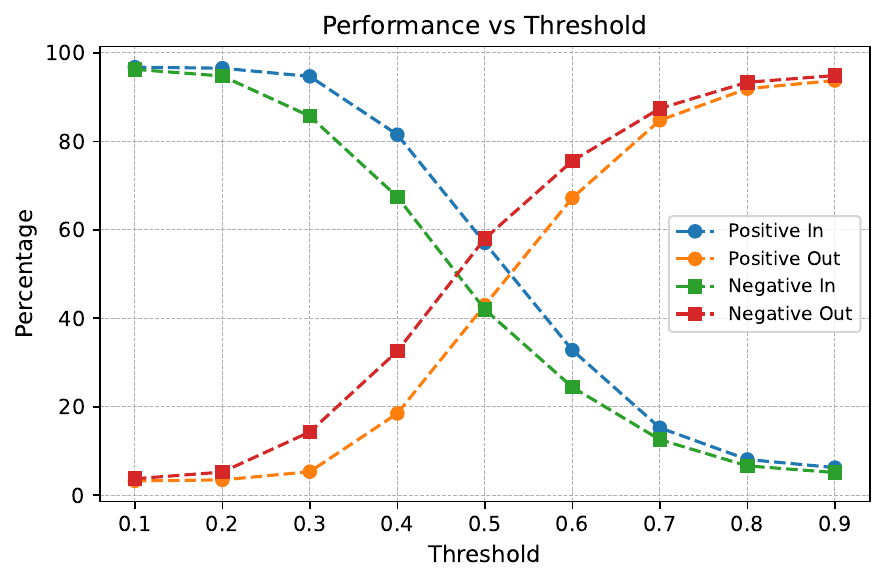}
}
\caption{Performance of LLM-generated homographic pun words interpretation under varying semantic similarity thresholds with \textit{WordNet}. The horizontal axis represents the semantic similarity threshold, and the vertical axis represents the number of samples with similarity greater than the threshold.}
\label{performance1}
\end{figure}

As shown in the formula, we evaluate the model's generated results by combining definitions from the open-source dictionary, specifically by calculating the cosine similarity score $S_{sem}(\cdot)$ between the generated definition and the dictionary definition.

\begin{equation}
S_{sem}(s_i,d_j)
=
\cos \big( e(s_i), e(d_j) \big)
\end{equation}
\begin{equation}
M(s_i)
=
\max_{d_j \in D(p_w)}
S_{sem}(s_i,d_j)
\end{equation}
\begin{equation}
\hat y=
\mathbb I
\left(
M(s_1)\ge\tau_{sem}
\land
M(s_2)\ge\tau_{sem}
\right)
\end{equation}
where $M(s_i)$ denotes the maximum matching similarity between a model-generated sense interpretation and the corresponding dictionary definitions. The model is considered to have correctly generated the two meanings of the pun only when both $M(s_1)$ and $M(s_2)$ exceed the predefined threshold. In addition, we define the proportion of positive samples $P_{in}(\tau)$ that successfully match the dictionary definitions, as formulated below:
\begin{equation}
P_{in}(\tau)
=
\frac{
\sum_{i \in P} \hat{y}_i
}{
|P|
}
\end{equation}
\begin{equation}
P_{out}(\tau) = 1 - P_{in}(\tau)
\end{equation}
And the proportions of negative samples $N_{in}(\tau)$: 
\begin{equation}
N_{in}(\tau)
=
\frac{
\sum_{i \in N} \hat{y}_i
}{
|N|
}
\end{equation}
\begin{equation}
N_{out}(\tau) = 1 - N_{in}(\tau)
\end{equation}
where $P_{out}(\tau)$ and $N_{out}(\tau)$ denote the proportions of positive and negative samples, respectively, whose generated sense explanations fail to match the dictionary definitions. 

As shown in Figure \ref{performance1}, within the dictionary definition space, correct predictions consistently achieve higher matching scores than incorrect predictions. This indicates that erroneous predictions are less likely to align with the dictionary-defined senses of the target pun word, suggesting that the generated sense explanations often deviate from the intended semantic space, so we thus have:

\noindent \textbf{Limitation 2: }LLMs are prone to uncontrolled generation during reasoning, resulting in generated sense explanations that fail to establish meaningful semantic associations with the original pun words.

\begin{figure*}[h] 
\centering
\resizebox{2\columnwidth}{!}{
\includegraphics[width=2.5\columnwidth,height=1.1\columnwidth]{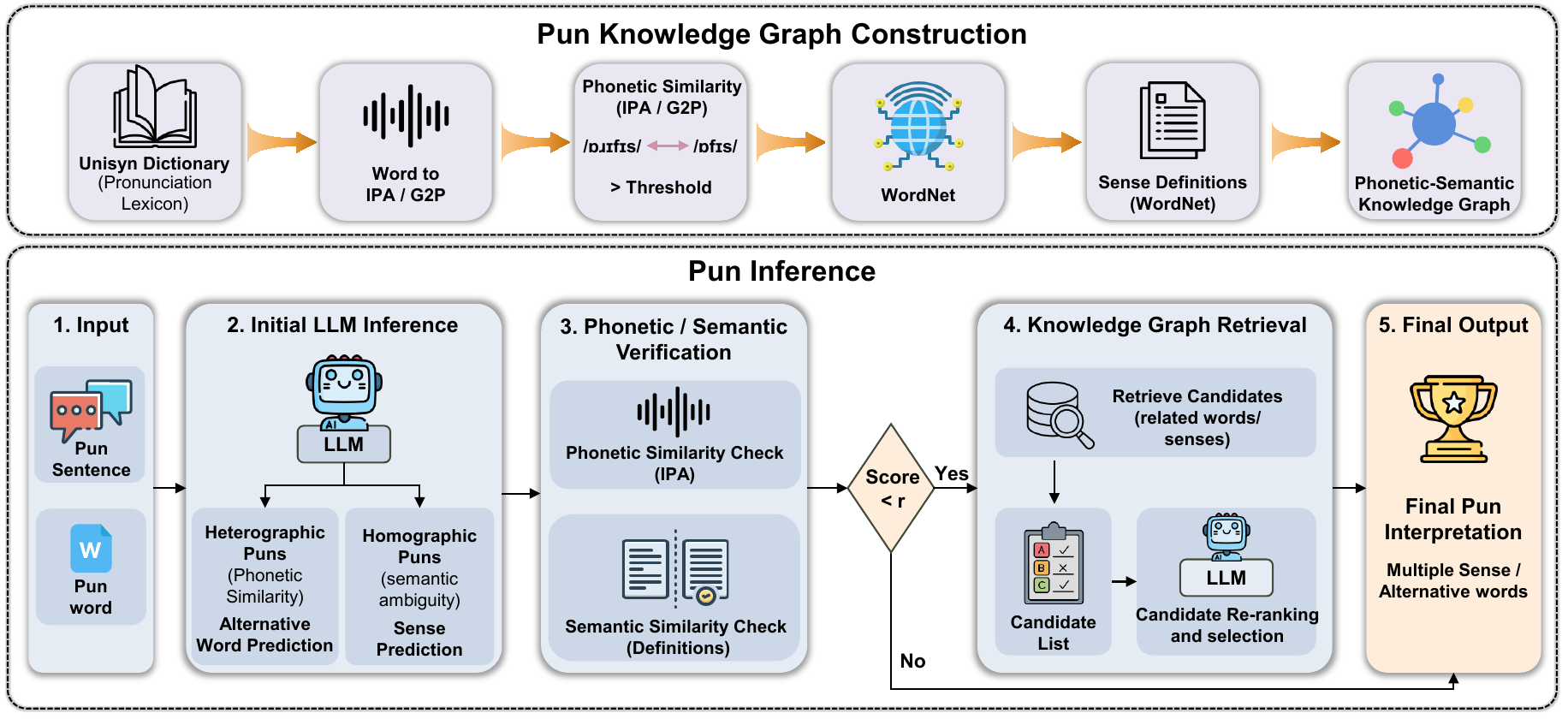}
}
\caption{The Overview of PunGraph Framework}
\label{overview}
\end{figure*}

In summary, the underperformance of small-scale LLMs in pun comprehension primarily stems from phonological reasoning deficiency and unconstrained semantic generation, which are two core limitations that we address in PunGraph.





\section{PunGraph}
This section introduces PunGraph, a framework specifically designed for pun reasoning that addresses the two core limitations identified above: insufficient phonological reasoning and unconstrained semantic generation. PunGraph constructs a lexical-level knowledge graph from a pronunciation dictionary to model phonological relationships between pun words, and further integrates dictionary-based sense definitions to retrieve candidate interpretations, providing LLMs with a constrained reasoning space, as illustrated in Figure \ref{overview}. Through this retrieval-enhanced reasoning process, PunGraph alleviates the weaknesses of small-scale LLMs in phonological modeling and semantic control, improving both the accuracy and controllability of pun reasoning.

\subsection{Pun Graph Construction}
To construct the knowledge graph, Inspired by \citep{manurung2008adding}, we generate word triples based on phonetic similarity relations derived from the Unisyn phonetic dictionary \cite{fitt1999synthesis}. We further incorporate IPA-based similarity mod-
eling and grapheme-to-phoneme (G2P) \cite{bisani2008joint} augmentation to enhance phonetic associations between words, building upon the original phonetic representations provided by the dictionary. Specifically, we compute the cosine similarity between the IPA representations of word pairs and establish connections between pairs whose similarity scores exceed a threshold, thereby constructing triples in the standard form of {source, relation, target}. Furthermore, we integrate the large-scale English lexical dictionary WordNet \citep{miller1995wordnet} to supplement semantic information for words that can be matched within the graph. The formulation of the knowledge graph is defined as follows:
\begin{equation}
\resizebox{0.7\hsize}{!}{$
\begin{aligned}
\mathcal{G} = \{\, &(w_i, r_{\text{phon}}, w_j,
  \mathcal{S}(w_i), \mathcal{S}(w_j))
  \mid \\
&e\bigl(\phi(w_i),\phi(w_j)\bigr)>0.75
\,\}
\end{aligned}
$}
\end{equation}
where $w_i$ and $w_j$ denote word entities, $r_{\text{phon}}$ represents the phonetic similarity relation, and $\mathcal{S}(w_i)$ and $\mathcal{S}(w_j)$ denote the corresponding semantic interpretations derived from the dictionary. The function $e(\cdot)$ denotes the phonetic similarity computation, with implementation details provided in Appendix \ref{Phonetic Similarity}. In addition, we import the constructed graph into the Neo4j graph database \footnote{https://neo4j.com/} for storage and visualization, thereby facilitating subsequent structural analysis and retrieval operations.

\subsection{Pun reasoning}
\label{Pun reasoning}
After constructing the knowledge graph, we design the corresponding reasoning tasks for heterographic and homographic puns, respectively. For each sample, given a pun $p$ and its pun words $p_w$, we first use prompts to guide the LLMs for preliminary reasoning to get alternative words $a_w$ and interpretation $[s_1,s_2]$, and then classify the results into two categories: correct predictions and incorrect predictions. For incorrect cases, we further introduce a graph retrieval enhancement mechanism to utilize external structured knowledge to assist the model in completing subsequent reasoning.

\textbf{Heterographic Puns} In this stage, we introduce a reasoning enhancement mechanism for heterographic puns. Specifically, as shown in Figure \ref{performance}, based on the performance trend of IPA similarity, we set 0.6 as the key threshold. When the similarity is below this threshold, the proportion of incorrect predictions increases significantly, indicating that the model has difficulty generating reasonable replacement words that are phonetically consistent with the pun within this range. Therefore, we focus on screening and processing samples with IPA similarity below 0.6. 

We retrieve one-hop adjacent near-homophones and semantically related words from the phonetic-semantic knowledge graph given a pun word  and construct a constrained candidate set $[a_1,a_2,a_3...,a_n]$. Specifically, the retrieval process traverses pronunciation similarity links and semantic association edges in the graph, where the average number of retrieved candidates is reported using the Candidate Retrieval Number (CRN) described in Appendix \ref{Data Statistics}. The retrieved candidates are subsequently formatted as explicit answer options and incorporated into the prompt, enabling the LLM to perform reasoning within a constrained candidate space and select the candidate that best fits the contextual semantics. Detailed prompt templates and candidate formatting strategies are provided in Appendix \ref{Retrieval Strategy}.

\textbf{Homographic Puns} We also consider the reasoning task for homographic puns. In the initial classification stage, we set the similarity threshold to 0.5, as Figure \ref{performance1} shows that this value yields the largest gap between positive and negative cases in terms of dictionary coverage. Specifically, when the similarity exceeds 0.5, a larger proportion of positive instances fall within the dictionary-defined sense space, whereas when the similarity is below 0.5, negative instances are more likely to fall outside this space. Based on this observation, we select error cases with similarity below 0.5 for subsequent knowledge graph enhancement.

Given a pun word $p_w$ we retrieve its associated entity attributes from the knowledge graph to obtain a set of candidate senses $[s_1,s_2]$, which are then formulated as multiple-choice options. Similar to the heterographic puns, we guide LLMs to select the most appropriate option within this constrained candidate space, thereby producing the final sense prediction for the pun word.

%

\section{WebPun Dataset}
We proposed a new benchmark to address the scarcity of high-quality data in pun reasoning.

\subsection{Data Preparation}
Existing work on pun understanding mainly relies on the SemEval-2017 benchmark, which is limited in scale and diversity, making it insufficient for evaluating modern LLMs on pun reasoning. To address this gap, we construct WebPun which is a new pun dataset collected from public pun websites, including Pun.me\footnote{https://pun.me/} and Punpedia\footnote{https://punpedia.org/}, and collect a total of 24,880 samples. We filter the dataset to retain pun sentences containing more than five words, while removing non-English samples and entries that do not form complete sentences. In addition, we remove rare pun words, as well as phrasal puns. Ultimately, a total of 5,730 samples are retained in the final dataset.

Since some pun sentences do not provide annotations for heterographic and homographic categories, we design a novel automated classification method. Specifically, we first employ the latest closed-source large model, GPT-5.5 \cite{openai2026gpt55}, to perform initial classification. To further improve annotation accuracy, we additionally introduce Gemini-3.5-flash \cite{googledeepmind2025gemini3flash} and Claude-Opus-4-1 \cite{claude_systemcard_2025} as auxiliary classification models and conduct cross-comparisons among the outputs of the three LLMs. When inconsistencies arise across model predictions, the corresponding samples are further submitted for manual review to determine the final classification labels. Further data annotation and classification prompt details can be shown at Appendix \ref{Classification Prompts}.

\subsection{Data Analysis}
\begin{figure}[t] 
\resizebox{\columnwidth}{!}{
\captionsetup{skip=6pt}
\includegraphics[width=\columnwidth,height=0.45\columnwidth, trim=0 0 0 0, clip]{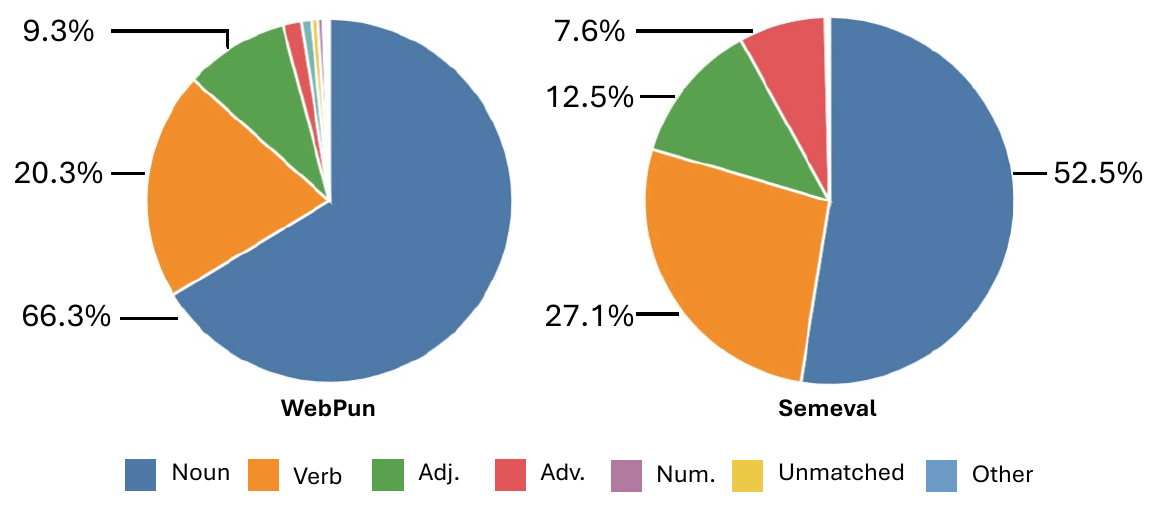}
}
\caption{Part-of-speech distribution of pun words in the WebPun and SemEval datasets.}
\label{data_analysis}
\end{figure}

After completing the annotation process, we further analyze the statistical characteristics of the dataset, as presented in Appendix \ref{Data Statistics}. WebPun contains 5,730 annotated pun instances, including 5,061 heterographic puns and 679 homographic puns. In addition, we conduct a part-of-speech analysis of the pun words in WebPun and compared the results with those of the SemEval dataset, as illustrated in Figure \ref{data_analysis}.

The results show that nouns account for more than half of the puns in the WebPun dataset, followed by verbs and adjectives. Together, these three categories constitute 95.9\% of all puns in WebPun. This distribution is consistent with that of the SemEval-2017 dataset, where the proportion of nouns is 92.1\%, suggesting that puns rely primarily on nouns, verbs, and adjectives to convey semantic ambiguity and humorous effects. Meanwhile, compared with existing pun datasets, WebPun further strengthens the coverage of the most common noun-based pun category, resulting in richer lexical diversity and semantic ambiguity. This characteristic makes WebPun a more challenging and effective benchmark for evaluating the pun understanding and reasoning capabilities of LLMs.

\begin{table*}[t]
\centering
\resizebox{2\columnwidth}{!}{
\begin{tabular}{lcccccccccccc}
\toprule
\multirow{3}{*}{\textbf{Model}} & 
\multicolumn{6}{c}{\textbf{Semeval 2017 Dataset}} &
\multicolumn{6}{c}{\textbf{WebPun Dataset}} \\
 & \multicolumn{3}{c}{\textbf{Heterographic Puns}} & 
\multicolumn{3}{c}{\textbf{Homographic Puns}} &
\multicolumn{3}{c}{\textbf{Heterographic Puns}} &
\multicolumn{3}{c}{\textbf{Homographic Puns}} \\ 
\cmidrule(lr){2-4}  \cmidrule(lr){5-7} \cmidrule(lr){8-10} \cmidrule(lr){11-13}
 & Pre ($\uparrow$) & Rec ($\uparrow$) & F1 ($\uparrow$) & Acc ($\uparrow$) & PMA ($\uparrow$) & F1 ($\uparrow$) & Pre ($\uparrow$) & Rec ($\uparrow$) & F1 ($\uparrow$) & Acc ($\uparrow$) & PMA ($\uparrow$) & F1 ($\uparrow$)\\
\midrule
\textit{\textbf{Large-scale LLMs}} \\
GPT-4o \cite{hurst2024gpt} & 76.95 &82.11&79.45 & 76.27 &98.54 &87.35 & 86.86 & 88.39 & 87.62 &70.85 &98.65 &84.75\\
Gemini-2.0 Flash \cite{gemini20flash2025} & 72.67 & 82.69 & 77.36 &71.08 &98.69 &84.56 & 61.26 &69.78 &65.24&68.74 &98.51 &83.06 \\
DeepSeek-V3.2 \cite{deepseekai2025deepseekv32} & 76.68 &84.28 &80.31 &66.26 &98.15 &82.12 &67.86 &75.01 &71.26 &66.77 &97.31 &82.02 \\
\addlinespace[0.5ex]   
\hdashline
\addlinespace[0.5ex]   
\textit{\textbf{Small-scale LLMs}} \\
MiniCPM-8.7B \cite{hu2024minicpm} & 31.32 &47.84 & 37.86&40.71 &93.07 &66.64 &22.20 & 30.63 & 25.74 &40.51 &94.92 &67.61\\
Qwen-2.5-7B \cite{qwen2.5} & 35.06 & 40.22&37.47 &34.65 &90.80 &62.42 & 30.33 & 35.27 & 32.61 &36.83 &92.07 &64.42\\
Qwen-3.5-27B \cite{qwen3.5} & 69.31 &79.51&74.06& 68.95 & 97.61 & 83.20 & 63.86 & 71.97 & 67.68&\underline{69.66} & 96.71& \underline{83.18} \\
Llama4-Maverick \cite{meta2025llama4} & 69.94 & 78.04 &73.77 &66.26 &\underline{97.84} & 82.00 & \underline{72.06} & \underline{76.88} & \underline{74.39} &67.56 &97.61 &82.59\\
\addlinespace[0.5ex]   
\hdashline
\addlinespace[0.5ex]   
\textit{\textbf{Specialized Reasoning Methods}} \\
PunIntended \cite{zeng2024barking} & 15.84 & 17.54 & 16.65 & 26.35 & 85.25 & 50.98 & 13.73 & 14.87 & 14.27 &29.66 &83.63 &51.60\\
GCR \cite{luo2024graph} &40.89 &68.44 &51.19 & 43.04 & 93.12 & 37.19 &36.86  &42.65 &39.54 &45.32 &94.26 &39.74\\
ReKG-MCTS \cite{song2025rekg} & 61.74 & 78.11 & 68.97 & 22.11 &78.04 &48.41 & 57.46 &72.61 & 64.15 & 21.46 & 82.93 & 52.17 \\
\addlinespace[0.5ex]   
\hdashline
\addlinespace[0.5ex]   
\textit{\textbf{Ours}} \\
\textbf{PunGraph-Qwen-2.5-7B} & 53.68 & 65.94 & 59.18 & 45.71 & 93.38 & 65.11 & 38.53 & 47.49&42.54 &45.44 &95.81 &67.45\\
\textbf{PunGraph-Qwen-3.5-27B} & \underline{74.40} & \underline{86.18} & \underline{79.86} & \textbf{76.18} &\textbf{98.84} &\textbf{85.71} &65.66 &76.06& 70.48& \textbf{78.33} & \underline{98.06} &\textbf{86.07}\\
\textbf{PunGraph-Llama4-Maverick} &\textbf{78.48} &\textbf{88.99} &\textbf{83.41} & \underline{71.80} & 97.46 & \underline{83.43} & \textbf{75.86} & \textbf{83.43} & \textbf{79.47} & 69.51 &\textbf{98.36} &82.62 \\

\bottomrule
\end{tabular}
}
\caption{Results of pun reasoning on the SemEval and WebPun datasets. Boldface indicates the best performance, while underlined values indicate the second-best performance among different methods. The models are categorized into Large-scale LLMs, Small-scale LLMs, Specialized Reasoning Methods, and Ours. Pre., Rec., F1, Acc., and PMA denote the evaluation metrics of precision, recall, F1-score, accuracy and partial matching accuracy, respectively.}
\label{main experiments}
\end{table*}

\section{Experiments}

\subsection{Dataset} To verify the effectiveness of our method, we conducted experiments on SemEval 2017 Dataset \cite{DBLP:journals/corr/AugensteinDRVM17} and WebPun, respectively. Table \ref{WebPun and Semeval} shows the data statistics of the two datasets. For the phonetic-semantic knowledge graph, we construct a total of 46,208 word entities and 760,312 phonetic similarity relations. We empirically set the threshold for establishing phonetic similarity links to 0.75.

\begin{table}[t]
\centering
\resizebox{0.7\columnwidth}{!}{
\begin{tabular}{lcc}
\toprule
\textbf{Model} & \textbf{Semeval} & \textbf{WebPun}\\
\midrule
Homographic & 1,298 &668\\
Heterographic  &1,098&5,061 \\
\bottomrule
\end{tabular}
}
\caption{The statistics of Semeval-2017 dataset and WebPun dataset.}
\label{WebPun and Semeval}
\end{table}

\subsection{Baselines} 
We select a diverse set of baseline models for comparison with our method, which are categorized into three groups: large-scale LLMs, small-scale LLMs, and specialized reasoning methods. 

The large-scale LLMs include GPT-4o \cite{hurst2024gpt}, Gemini 2.0 Flash \cite{gemini20flash2025}, and DeepSeek-V3.2 \cite{deepseekai2025deepseekv32}. The small-scale LLMs consist of MiniCPM-8.7B \cite{hu2024minicpm}, Qwen-2.5-7B, Qwen-2.5-70B \cite{qwen2.5}, Qwen-3.5-27B \cite{qwen3.5} and Llama4-maverick \cite{meta2025llama4}. For specialized reasoning methods, we reproduce the PunIntended framework \cite{zeng2024barking}, while also considering the graph-based reasoning methods, GCR \cite{luo2024graph} and ReKG-MCTS \cite{song2025rekg}, for further comparison.

\subsection{Evaluation Metrics}
We conduct separate evaluations for heterographic and homographic puns. For heterographic puns, we first normalize word forms to reduce the impact of surface-level variations such as capitalization, singular/plural inflections, and tense changes. Inspired by \cite{su2026words}, we perform exact matching on the alternative words predicted by the model and adopt precision, recall, and F1 score as evaluation metrics. For homographic puns, we compute pairwise semantic similarity between the gold-standard sense explanations and the sense explanations generated by the model on both the SemEval 2017 Dataset \cite{DBLP:journals/corr/AugensteinDRVM17,su2026words} and WebPun, and use accuracy and F1 score to evaluate the overall performance. A prediction is regarded as correct only when the similarity scores of the corresponding sense explanations both exceed a predefined threshold. In addition, we further define a Partial Matching Accuracy (PMA) metric, which is counted as correct under PMA when at least one of the two generated sense explanations achieves a similarity score higher than the predefined threshold with the corresponding gold-standard explanation.

\begin{table}[t]
\centering
\resizebox{\columnwidth}{!}{
\begin{tabular}{lcccc}
\toprule
\textbf{Model} & \textbf{SemEval} & \textbf{SemEval} & \textbf{WebPun} & \textbf{WebPun} \\
 & \textbf{Het.} & \textbf{Hom.} & \textbf{Het.} & \textbf{Hom.} \\
\midrule
Qwen-2.5-7B & +57.95\% & +4.31\% & +30.45\% & +4.70\% \\
Qwen-3.5-27B & +7.83\% & +3.02\% & +4.14\% & +3.47\% \\
Llama4-Maverick & +13.06\% & +1.74\% & +6.83\% & +0.04\% \\
\bottomrule
\end{tabular}
}
\caption{Relative F1-score improvement of PunGraph over the corresponding small-scale LLMs across different datasets and pun types. Het. and Hom. denote heterographic and homographic puns, respectively.}
\label{tab:relative_improvement}
\end{table}

\section{Results and Discussion}

\subsection{Main Results}
Table \ref{main experiments} presents the comparative results of PunGraph against different categories of baseline methods and table \ref{tab:relative_improvement} reports the relative F1-score improvement of backbone small-scale LLMs. First, compared with direct reasoning using the same backbone models, PunGraph consistently improves performance across both datasets and both pun reasoning tasks, demonstrating the effectiveness of retrieval-enhanced structured knowledge augmentation. For example, with Qwen-2.5-7B, PunGraph achieves relative F1 improvements of 57.95\% on SemEval and 30.45\% on WebPun for heterographic puns. For homographic puns, it further yields gains of 4.31\% and 4.70\%, respectively. Similar improvements are observed on Qwen-3.5-27B and Llama4-Maverick, indicating that PunGraph is not tied to a specific backbone model, but serves as a general and effective external knowledge augmentation framework for pun reasoning. In addition, PunGraph remains highly competitive when compared with large-scale proprietary LLMs. Notably, on the SemEval heterographic pun task, PunGraph-Llama4-Maverick achieves an F1 score of 83.41, outperforming GPT-4o, Gemini-2.0 Flash, and DeepSeek-V3.2. These results suggest explicitly modeling phonological similarity and constraining reasoning within a structured semantic candidate space can effectively compensate for the limitations of smaller models in data coverage, implicit phonological reasoning, and controllable generation.

Finally, PunGraph consistently outperforms specialized reasoning baselines across both datasets. On SemEval, PunGraph-Llama4-Maverick surpasses ReKG-MCTS by 14.44 F1 on heterographic puns, while on WebPun it further achieves gains of +15.32 and +33.90 F1 on heterographic and homographic puns, respectively. These results highlight the advantage of PunGraph in modeling phonological-semantic interactions through task-specific retrieval-enhanced reasoning.



\subsection{Ablation Study}



To evaluate the effectiveness of different pronunciation link construction methods in the knowledge graph, we conduct an ablation study to compare the impact of different phonetic connection strategies on model performance. Specifically, (1) w/o IPA removes only the IPA-based similarity modeling relations; (2) w/o G2P removes only the pronunciation association relations introduced through G2P augmentation; and (3) w/o Unisyn removes the similarity links constructed from the Unisyn pronunciation dictionary. Through these experiments, we aim to investigate the contribution of different phonetic modeling methods to the construction of pronunciation-aware graph connections and their influence on pun reasoning performance.

Table \ref{phonetic symbols} reports the ablation results for different pronunciation knowledge components in PunGraph. The results demonstrate that integrating multiple pronunciation modeling strategies consistently improves performance across datasets and backbone models. For instance, on the SemEval-2017 dataset with Qwen2.5-7B as the backbone, removing IPA-based similarity modeling, G2P augmentation, and the Unisyn pronunciation dictionary leads to F1-score drops of 8.61\%, 5.19\%, and 12.42\%, respectively. Notably, across both Qwen2.5-7B and Llama4-Maverick backbones and on both the SemEval-2017 and WebPun datasets, removing G2P consistently results in the smallest performance degradation, whereas removing the Unisyn pronunciation dictionary causes the most substantial decline. These findings suggest that pronunciation similarity relations derived from Unisyn play a central role in PunGraph’s pronunciation association modeling, while G2P augmentation primarily serves as a complementary mechanism that enhances pronunciation coverage and improves the robustness of the constructed graph.

\begin{table}[t]
\centering
\resizebox{\columnwidth}{!}{
\begin{tabular}{lcccccc}
\toprule
\multirow{2}{*}{\textbf{Model}} & 
\multicolumn{3}{c}{\textbf{SemEval 2017 Dataset}} &
\multicolumn{3}{c}{\textbf{WebPun Dataset}} \\
 & Pre ($\uparrow$) & Rec ($\uparrow$) & F1 ($\uparrow$) & Pre ($\uparrow$) & Rec ($\uparrow$) & F1 ($\uparrow$)\\
\midrule
Qwen2.5-7B\\
PunGraph & \textbf{53.68}& \textbf{65.94} &\textbf{59.18} & \textbf{38.53}& \textbf{47.49}& \textbf{42.54} \\
w/o IPA  &46.28&55.73 & 50.57 &29.60 &37.69 &33.16 \\
w/o G2P  &48.78&60.46 & 53.99 &29.73 &38.82 &33.67 \\
w/o Unisyn  &42.58&51.85 &46.76 &26.20 &32.67 &29.07 \\
\hdashline
Llama4-Maverick\\
PunGraph & \textbf{78.48}& \textbf{88.99}& \textbf{83.41}&\textbf{75.86}& \textbf{83.43}& \textbf{79.47}\\
w/o IPA  &71.49&82.98 & 76.81 &73.26&81.34&77.09 \\
w/o G2P  &72.67&83.82 & 77.85 &73.40&81.25&77.12\\
w/o Unisyn  &70.67&80.58 & 75.30 &70.80&77.80&74.13\\
\bottomrule
\end{tabular}
}
\caption{Ablation results of different phonetic knowledge components in PunGraph using Qwen2.5-7B and Llama4-Maverick, including IPA-based similarity modeling, G2P augmentation, and the Unisyn pronunciation dictionary.}
\label{phonetic symbols}
\end{table}

\subsection{Error Analysis}
We conduct an error analysis to investigate the performance of the proposed method on the pun reasoning task and categorize the error types for both homographic and heterographic puns. Specifically, the errors can be grouped into three main categories: (1) missing phonetic similarity links in the graph or the failure of the dictionary to provide the corresponding definitions, which prevents the model from retrieving valid word or definition candidates; (2) incorrect selection among the retrieved candidate words or definition options; and (3) generation errors produced by the model itself, such as failing to follow the prompt instructions. We present the corresponding error statistics for Qwen-2.5-7B as a representative example according to homographic and heterographic puns.

\begin{figure}[t] 
\resizebox{\columnwidth}{!}{
\captionsetup{skip=6pt}
\includegraphics[width=\columnwidth,height=0.55\columnwidth, trim=0 0 0 0, clip]{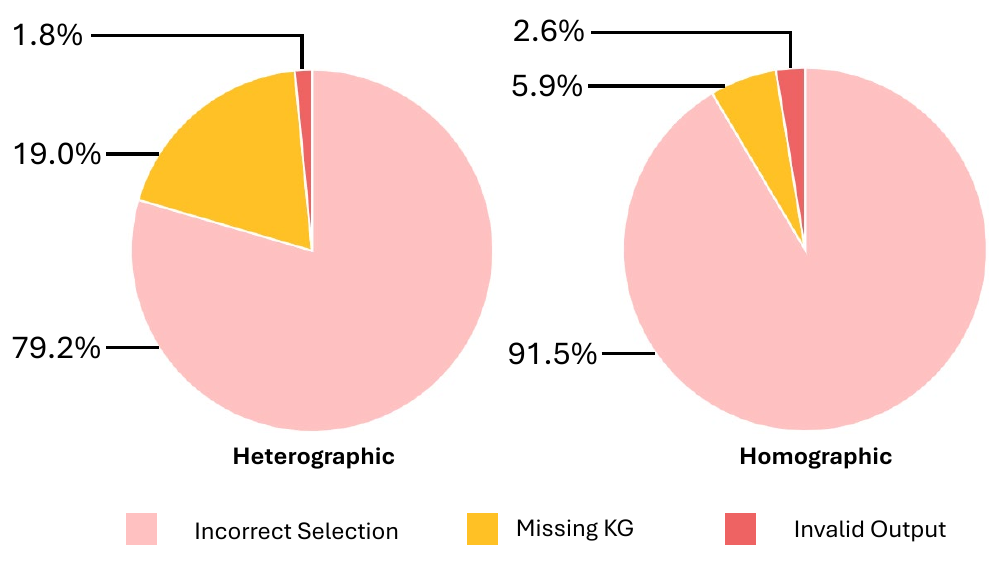}
}
\caption{Error cases of heterographic and homographic puns in the SemEval-2017 dataset. Selection Error, Missing KG, and Invalid Output denote cases where LLMs select incorrect candidates, the knowledge graph retrieval misses the ground-truth answer, and the model fails to follow the required output format, respectively.}
\label{Case Study}
\end{figure}
As shown in the Figure \ref{Case Study}, the primary source of errors in our method is the model’s tendency to confuse the correct answer with other candidate options, thereby leading to reasoning failures. This issue is particularly pronounced for homographic puns, where such errors account for 91.5\% of all failures. In contrast, errors caused by knowledge graph retrieval account for only 19\% and 5.9\% in heterographic and homographic puns, respectively. These results indicate that the constructed phonetic similarity links and dictionary definitions provide effective coverage of the pun datasets.

\section{Related Works}

\subsection{Pun Interpretation}

Existing work on pun interpretation mainly follows two directions: semantics-based methods and pronunciation-aware methods. SemEval-2017 Task 7 formalized English pun processing into detection, location, and interpretation subtasks, providing a standard benchmark for later studies \cite{miller2017semeval}. Subsequent work extended this setting to joint detection and location, multilingual pun interpretation, and task-specific pun modeling \cite{zou2019joint,prnjak2023clef,chen2024u}. Semantics-based methods use lexical resources such as WordNet or distributed representations to model word senses and semantic relatedness \cite{miller1995wordnet,miller2017semeval,zhou2020boating}. They are suitable for homographic puns, but are less reliable for heterographic puns that depend on latent pronunciation-based alternatives. Pronunciation-aware methods address this issue by using pronunciation dictionaries, phonological resources, phoneme-level representations, or pronunciation-aware attention mechanisms \cite{manurung2008adding,zhou2020boating}. Pun generation studies further highlight the interaction among the pun word, the alternative word, and context \cite{yu2018neural,he2019pun,sun2022context,mittal2022ambipun,tian2022unified}. However, semantic and phonological methods operate in different spaces, and LLMs may still miss alternative words or phonetic links \cite{xu2024good,zangari2025pun}.

\subsection{Retrieval-Augmented Reasoning}

Retrieval-augmented generation (RAG) enhances LLMs by combining parametric model memory with non-parametric knowledge retrieved from external corpora, improving knowledge access and generation quality in knowledge-intensive tasks \cite{lewis2020retrieval}. Conventional RAG mainly retrieves unstructured text passages, while GraphRAG introduces structured graph information, such as nodes, triples, paths, or subgraphs, into retrieval and generation \cite{peng2024graphrag}. In knowledge graph reasoning, recent work retrieves graph paths or subgraphs as evidence for LLM reasoning \cite{luo2024graph,song2025rekg,liu-etal-2025-david}; GNN-RAG further uses graph neural networks to score candidate answers and retrieve connecting paths \cite{mavromatis2025gnnrag}. These methods mainly focus on factual or entity-relation reasoning, rather than the joint phonological-semantic structure required for pun interpretation. PunGraph adapts retrieval-augmented reasoning to this setting by retrieving candidate words or senses from a task-specific lexical graph.

\section{Conclusion}
In this paper, we proposed PunGraph, a retrieval-enhanced knowledge graph framework for pun understanding. By integrating phonetic similarity and semantic knowledge into a structured lexical graph, PunGraph enables LLMs to perform more controllable and accurate reasoning for both heterographic and homographic puns. We further introduced WebPun, a new large-scale pun reasoning dataset containing 5,730 annotated samples collected from online pun resources. Experimental results on SemEval-2017 and WebPun demonstrate that PunGraph consistently improves the performance of small-scale LLMs and achieves competitive results compared with strong proprietary models. Further analysis shows that retrieval-guided phonetic and semantic constraints effectively reduce common errors in pun reasoning. 
\section*{Limitations}
Despite the promising performance of PunGraph, our work has several limitations. First, our framework is designed primarily for English and relies on English-specific lexical resources such as Unisyn and WordNet, which may limit its direct applicability to other languages. Second, the effectiveness of PunGraph depends on the coverage of the constructed phonetic-semantic knowledge graph; rare words, creative expressions, or unseen pun patterns may not be well represented in the graph, leading to retrieval failures. In addition, although retrieval constrains the reasoning space of LLMs, the final prediction still depends on the model’s ability to select the correct candidate, and incorrect candidate selection remains a major source of errors. Finally, WebPun is collected from online written pun resources and may not fully capture broader forms of humor such as spoken, multimodal, or culturally specific puns. We leave multilingual and multimodal extensions of PunGraph for future work.
\section*{Ethical Considerations}
The human-participant study presented in this paper was approved by the appropriate ethics committee of Shanxi University. All procedures involving human participants were conducted in accordance with the relevant ethical guidelines and regulations.
\section*{Acknowledgments}
This research is supported by the Strong AI Lab and the Natural, Artificial, and Organisation Intelligence Institute at the University of Auckland. The first author of this research is funded by the China Scholarship Council (CSC). 


\bibliography{custom}

\appendix

\section{Annotation Details}

\subsection{Classification Prompts}
\label{Classification Prompts}
We report the prompt of LLMs for classifying pun types, as follows:
\begin{tcolorbox}[title=Prompt for Pun Type Classification]
\small
\textbf{System Prompt:}

You are an expert in English pun analysis.

Your task is to classify the following sentence as either a homographic pun or a heterographic pun.

\textbf{Definitions:}

1. Homographic pun:  
A pun in which the same written word carries two or more different meanings within the sentence.

Example:  
``I used to be a banker, but I lost interest.''  
$\rightarrow$ ``interest'' has multiple meanings.

2. Heterographic pun:  
A pun in which two different words or phrases have different spellings but similar pronunciation, creating humorous ambiguity.

Example:  
``Dentists don't like a hard day at the orifice.''  
$\rightarrow$ ``orifice'' sounds similar to ``office''.

\textbf{Instructions:}

Read the sentence carefully. Identify the main pun word. Determine whether the humor is created primarily through semantic ambiguity of the same word, corresponding to a homographic pun, or phonetic similarity between different words, corresponding to a heterographic pun.

Return valid JSON only in the following format:

\begin{lstlisting}[breaklines=true]
{
  "pun_type": "homographic pun" or "heterographic pun"
}
\end{lstlisting}

\textbf{Rules:}

Return only one label. The pun word must be the exact word appearing in the sentence. Do not include explanations. Output JSON only.

\vspace{0.5em}
\textbf{User Input:}

sentence: \{sentence\}
\end{tcolorbox}

\begin{table*}[t]
\centering
\small
\resizebox{0.95\textwidth}{!}{
\begin{tabular}{p{6.3cm} p{1.8cm} p{2.2cm} p{7.5cm}}
\toprule
\textbf{Pun Sentence} & \textbf{Pun Word} & \textbf{Label} & \textbf{Reasoning} \\
\midrule

I used to be a banker, but I lost interest. 
& interest 
& Homographic 
& financial interest earned from money, personal interest or enthusiasm. \\

When I grow up I wanna be a cup so I can fight crime.
& cup
& Heterographic 
&  cop \\

Time flies like an arrow; fruit flies like a banana. 
& flies 
& Homographic 
& to move through the air, producing dual meanings in context. \\

The interrogators soon got a confection out of him.
& confection
& Heterographic
& confession \\

\bottomrule
\end{tabular}
}
\caption{Examples of annotated pun types in the WebPun dataset.}
\label{tab:annotation_examples}
\end{table*}

\subsection{Data Annotation}
Inspired by \cite{chen2024u}, we adopt a few-shot prompting strategy to perform preliminary annotation with large language models. Specifically, we first select three homographic and three heterographic pun examples from the SemEval dataset, and combine these examples together with their corresponding annotations and prompt instructions to guide the models in subsequent annotation tasks. Notably, for heterographic puns, the annotation target is the corresponding ground-truth replacement word, while for homographic puns, the annotation consists of the two sense interpretations of the pun word, uniformly represented in the format of $[s_1,s_2]$.

Similar to the pun type classification process, we also employ three closed-source large language models for collaborative annotation. Samples with consistent annotations across all three models are further reviewed by an expert annotator for quality assurance. In addition, we place particular emphasis on cases where the three models produce inconsistent annotations. For such samples, three expert annotators \footnote{The expert annotators were volunteers from diverse regional backgrounds, including former educated youth participants.} jointly conduct manual corrections, and the final annotation is determined through majority voting. Furthermore, we evaluate inter-annotator agreement by randomly sampling 150 annotated instances and measuring consistency using Fleiss’ Kappa \cite{fleiss1973equivalence}. The final agreement score is 0.54, indicating a moderate level of agreement among annotators and providing reasonable support for the consistency of the annotations.

\section{Dataset details}
\subsection{Data Statistics}
\label{Data Statistics}

We further conduct a statistical analysis of the SemEval and WebPun datasets, as shown in Table \ref{data statistic 2}. Specifically, we report the average sentence length, total number of words, and number of unique pun words for each dataset. The results show that although WebPun contains substantially more heterographic pun samples than SemEval, the increase in the number of unique pun words is relatively limited. This suggests a noticeable reuse of high-frequency pun words in real-world online puns, where multiple pun expressions are often constructed around the same or semantically related core pun words.

In addition, we analyze the number of candidate items retrieved for each pun word in PunGraph, denoted as the Candidate Retrieve Number (CRN). For heterographic puns, CRN refers to the number of candidate words retrieved through one-hop phonetic relation search in the graph; for homographic puns, it refers to the number of candidate sense definitions retrieved based on lexical semantic entries. The statistics show that PunGraph produces a relatively large retrieval space on the SemEval 2017 dataset, with an average of 30.55 candidate words for each heterographic pun and 9.76 candidate sense definitions for each homographic pun. It is worth noting that a larger candidate space does not necessarily lead to better reasoning performance, as excessive candidates may introduce additional noise and increase the difficulty of candidate selection. Therefore, CRN is mainly used to characterize the coverage of the graph retrieval space, rather than as a direct indicator of downstream reasoning performance.

\begin{table}[t]
\centering
\resizebox{\columnwidth}{!}{
\begin{tabular}{lccccccc}
\toprule
\textbf{Type} & Total & Sentence & Words & OnePun & CRN \\
\midrule
\textit{\textbf{SemEval 2017 Dataset}} \\
Heterographic & 1,098 & 11.79 & 12,949 &896 & 30.55 \\
Homographic  & 1,298 & 11.71 &15,194&928 & 9.76\\
\addlinespace[0.5ex]   
\hdashline
\addlinespace[0.5ex]   
\textit{\textbf{WebPun Dataset}} \\
Heterographic & 5,061 & 6.95 & 35,190&1,663 & 22.50 \\
Homographic  & 669 &11.98 &8,015&483 & 11.37\\
\bottomrule
\end{tabular}
}
\caption{Additional statistics of the SemEval-2017 and WebPun datasets. Total, Sentence, and Words denote the number of samples, average sentence length, and total number of words, respectively. OnePun refers to the number of unique pun words, while CRN denotes the Candidate Retrieve Number obtained from PunGraph.}
\label{data statistic 2}
\end{table}

\subsection{Dataset Sample}
Table~\ref{tab:annotation_examples} presents representative examples from WebPun with their corresponding pun words, types, and explanations.

\subsection{Dataset Novelty and Knowledge Coverage}

To further examine the novelty of WebPun with respect to existing pun benchmarks, we analyze the overlap of unique pun words between WebPun and SemEval-2017. Using SemEval-2017 as the denominator, the overlap is 18.99\% for heterographic puns and 15.52\% for homographic puns, as shown in Table~\ref{tab:pun_overlap}. The relatively limited overlap indicates that WebPun provides substantial complementary lexical coverage beyond SemEval-2017, despite the reuse of some high-frequency pun words across the two datasets.

\begin{table}[t]
\centering
\small
\resizebox{\columnwidth}{!}{
\begin{tabular}{lcc}
\toprule
\textbf{WebPun} 
& \textbf{SemEval Het. (895)} 
& \textbf{SemEval Hom. (928)} \\
\midrule
Heterographic 
& 170 (18.99\%) 
& 205 (22.09\%) \\
Homographic 
& 22 (2.46\%) 
& 144 (15.52\%) \\
\bottomrule
\end{tabular}
}
\caption{Overlap of unique pun words between WebPun and SemEval-2017. SemEval-2017 is used as the denominator. Values indicate the number and percentage of overlapping unique pun words.}
\label{tab:pun_overlap}
\end{table}

We further evaluate the coverage of the constructed phonetic-semantic knowledge graph on both datasets. As shown in Table~\ref{tab:kg_coverage}, the graph achieves 99.84\% and 85.95\% coverage for heterographic and homographic puns in WebPun, respectively, compared with 99.45\% and 86.67\% on SemEval-2017. Overall, the knowledge graph maintains consistently high coverage across both datasets, suggesting that the additional lexical diversity introduced by WebPun remains well supported by the phonetic and semantic resources used in PunGraph.

\begin{table}[t]
\centering
\resizebox{\columnwidth}{!}{
\begin{tabular}{lccc}
\toprule
Dataset & Heterographic & Homographic & Average \\
\midrule
WebPun & 99.84\% & 85.95\% & 98.22\% \\
SemEval-2017 & 99.45\% & 86.67\% & 92.53\% \\
\bottomrule
\end{tabular}
}
\caption{Knowledge-graph coverage on WebPun and SemEval-2017.}
\label{tab:kg_coverage}
\end{table}

Together, the relatively low overlap in unique pun words and the high knowledge-graph coverage suggest that WebPun complements SemEval-2017 with additional lexical and contextual diversity while remaining well supported by the structured knowledge used in PunGraph.

\section{Retrieval Strategy}
\label{Retrieval Strategy}
Regarding Section \ref{Pun reasoning}, we also provide the prompt template used to guide LLMs to process the retrieved candidate words according to the heterographic and homographic puns, as shown below:
\begin{tcolorbox}[title=Prompt for Heterographic Pun reasoning]
\small
\textbf{System Prompt:}

You are a linguist specializing in puns. Given a pun ``\{sentence\}'' and the pun word ``\{word\}'', please choose the most likely intended real word in the original pun sentence.

Output only the option word of dictionary, with no additional text.

If there is no suitable candidate, please generate the most likely real word based on your linguistic knowledge and the context of the sentence, without being limited to the candidate list.

\vspace{0.5em}
\textbf{User Input:}

Sentence: The key to changing your performance ability is by tuning out criticism and staying musically octave.

Pun word: octave

Choices: \{'artiste', 'octant', 'argive', 'octet', 'optics', 'arctic', 'octal', 'fictive', 'optic', 'octavo', 'active'\}
\end{tcolorbox}

\begin{tcolorbox}[title=Prompt for Homographic Pun Sense Selection]
\small
\textbf{System Prompt:}

You are an expert in multiple-choice question answering focused on explaining puns. Select the two most appropriate definition numbers that best explain the pun word in the given sentence.

If suitable definitions are available, respond only with an array of two option numbers, for example [1, 3].

If there are no suitable candidate definitions, infer its two distinct meanings and output two replacement words or phrases, each representing one meaning. In that case, provide only the two items, separated by a comma, with no additional text.

\vspace{0.5em}
\textbf{User Input:}

pun sentence: \{Old math profs never die, they just can't differentiate.\}

pun word: \{differentiate\}

definitions: \{1. the case expressing ownership;
2. serving to express or indicate possession;
3. desirous of owning;
4. having or showing a desire to control or dominate.\}
\end{tcolorbox}

\section{Phonetic Similarity}
\label{Phonetic Similarity}
We first compute edit distance based on the IPA representations of words to determine their phonetic similarity. However, phonetic similarity is not uniform across phonemes. For example, the vowel /i/ is phonetically closer to /e/ than to /u/. To better capture these graded phonetic relationships, we construct a phoneme similarity table based on articulatory phonetic features \footnote{https://www.internationalphoneticassociation.org/content/chart} following \cite{international1999handbook}, as shown in Table \ref{tab:phoneme_similarity}. This allows phoneme-level similarity to be modeled at a finer granularity. During edit distance computation, phonemes belonging to the same similarity group are treated as equivalent matches.
\begin{table}[t]
\centering
\small
\resizebox{\columnwidth}{!}{%
\begin{tabular}{lll}
\toprule
\textbf{Category} & \textbf{Phoneme Group} & \textbf{Similarity Basis} \\
\midrule
Consonant & \{p, b, m\} & bilabial; voicing / nasal variation \\
Consonant & \{t, d, n\} & alveolar; voicing / nasal variation \\
Consonant & \{s, z\} & alveolar fricatives; voicing contrast \\
Consonant & \{k, g\} & velar; voicing contrast \\
Consonant & \{f, v\} & labiodental fricatives; voicing contrast \\
Consonant & \{l, r\} & liquid consonants; approximant similarity \\
\midrule
Vowel & \{i, \textipa{I}\} & front vowels with adjacent height \\
Vowel & \{e, \textipa{E}, \textipa{\ae}\} &
front vowels with similar tongue height \\
Vowel & \{u, \textipa{U}\} &
back rounded vowels with adjacent height \\
Vowel & \{\textipa{O}, \textipa{A}\} &
back vowels with similar openness \\
Vowel & \{\textipa{@}, \textipa{V}\} &
central vowels with similar tongue position \\
Vowel & \{\textipa{O}, o\textipa{U}, u\} &
back rounded vowel cluster \\
\bottomrule
\end{tabular}%
}
\caption{Phonetically similar phoneme groups used for phonetic neighbor
retrieval in PunGraph. Phoneme similarity is modeled as an undirected
relation based on shared articulatory features.}
\label{tab:phoneme_similarity}
\end{table}

\end{document}